\documentclass{article}
\usepackage[utf8]{inputenc}
\usepackage[paper=letterpaper, margin=1in]{geometry}  
\usepackage{authblk}
\usepackage[numbers,sort&compress]{natbib}
\usepackage{comment}
\usepackage{booktabs}
\usepackage{tabularx}
\usepackage{bm}
\usepackage{hyperref}
\usepackage{graphicx}
\usepackage{xcolor}
\usepackage{subfig}

\graphicspath{ {./} }

\bibpunct{[}{]}{,}{n}{}{;}

\title{Grasp and Motion Planning for Dexterous Manipulation for the Real Robot Challenge
}   
\author[1]{Takuma Yoneda\thanks{Equal contribution}}
\author[1]{Charles Schaff\protect\footnotemark[1]}
\author[2]{Takahiro Maeda}
\author[1]{Matthew R.~Walter}

\affil[1]{Toyota Technological Institute at Chicago}
\affil[2]{Toyota Technological Institute}
\date{}

\begin{document}

\maketitle

\begin{abstract}
This report describes our winning submission to the \href{https://real-robot-challenge.com/}{Real Robot Challenge}.
The Real Robot Challenge is a three-phase dexterous manipulation competition that involves manipulating various rectangular objects with the TriFinger Platform.
Our approach combines motion planning with several motion primitives to manipulate the object. For Phases~$1$ and $2$, we additionally learn a residual policy in simulation that applies corrective actions on top of our controller.
Our approach won first place in Phase~$2$ and Phase~$3$ of the competition. We were anonymously known as `ardentstork' on the competition \href{https://real-robot-challenge.com/leader-board}{leaderboard}.
Videos and our code can be found at \href{https://github.com/ripl-ttic/real-robot-challenge}{https://github.com/ripl-ttic/real-robot-challenge}.  

\end{abstract}

\section{Real Robot Challenge}
The \href{https://real-robot-challenge.com/}{Real Robot Challenge} is a manipulation challenge using the TriFinger Platform developed by \citet{wuthrich2020trifinger}. The TriFinger robot consists of three identical fingers, each with three degrees of freedom, that are positioned above the robot's workspace. We are tasked with manipulating rectangular objects by moving them to various goal poses. The challenge consists of three phases. In Phase~$1$, the task is to manipulate a cube in a simulated environment. In Phase~$2$, the objective is to manipulate a cube using the real platform. Phase~$3$ increases the level of difficulty by requiring the manipulation of a much smaller cuboid object.

\begin{figure}[!t]%
    \centering
    \subfloat[\centering Phase~$2$: cube]{{\includegraphics[width=0.45\linewidth]{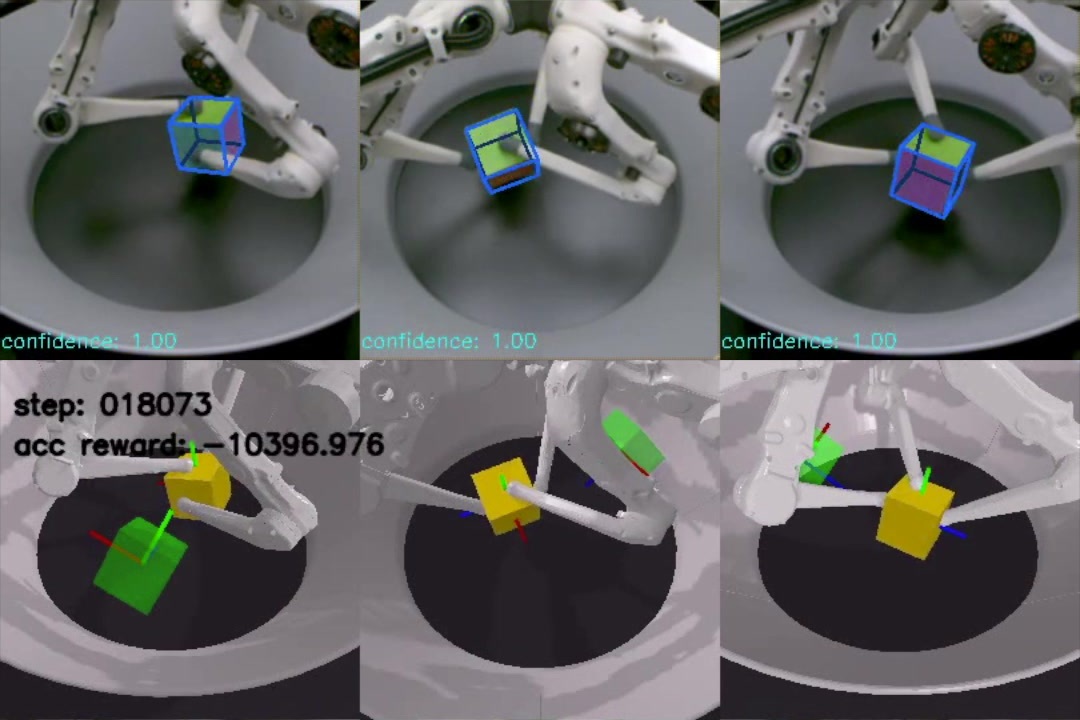} }}%
    \qquad
    \subfloat[\centering Phase~$3$: cuboid]{{\includegraphics[width=0.45\linewidth]{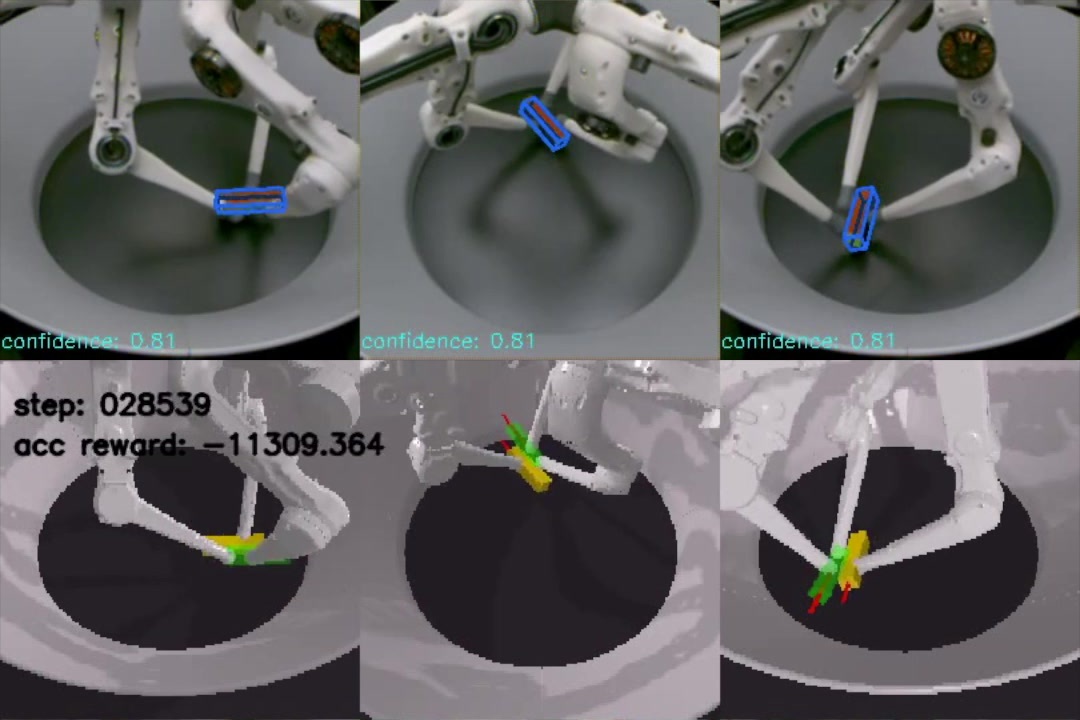} }}%
    \caption{Screenshots from (left) Phases $2$ and (right) Phase~$3$ at level~4. The three columns show views from cameras evenly distributed above the workspace as part of the single TriFinger Platform. The top row shows the views of the physical robot, while the bottom row visualizes a rendering of the environment reconstructed from proprioceptive (encoders) and exteroceptive (cameras) sensors. The goal pose is visualized in green.}
    \label{fig:screenshots}%
\end{figure}

In each phase, tasks are classified into four levels according to the goal pose of the object. In levels~$1-3$, the goal specifies the position of the object, but not the orientation. In level~$1$, the goal position is always on the ground; in level~$2$, it is fixed at the center and $20$\,cm above the ground plane; in level~$3$, it is randomly sampled within the robot's workspace; and in level~$4$, a goal orientation is additionally specified.\footnote{For tasks involving the cuboid object (Phase~$3$), it is not necessary to achieve a desired rotation around the long axis of the cuboid, due to the difficulty of accurately measuring it.} Performance is evaluated in terms of the speed and accuracy with which the goal is reached. At each timestep, a reward is provided based on the distance between the current object pose and the goal pose, in which positional and orientational differences are weighed differently and added up. The sum of reward over a $2$ minute period is used to evaluate each submission.
Figure~\ref{fig:screenshots} provides visualizations for Phase~$2$ and Phase~$3$.

The challenge organizer has provided an interface that estimates object pose from 3 camera views and returns the estimated pose as well as sensor readings of joint angles, velocities and tip force sensors on TriFinger Platform.
Participants are free to use their own algorithms to estimate object pose from the camera views, however, we used the provided estimation due to time limitation.
Further details regarding the challenge can be found at the \href{https://real-robot-challenge.com/}{official website}.

\section{Approach}
%
%
Achieving the goal pose often requires multiple grasps and manipulations. We find that many feasible grasps of the object are not suitable for actually solving the task, especially when matching orientation is required. 
Therefore, a successful approach must be able to search over the space of possible grasps. The small size of the object also requires precise positioning of the fingertips, which can be challenging in the presence of observation noise.

The core of our approach is a grasp-planning loop in which we sample feasible grasps until we find one for which a path exists that achieves the desired object pose. Here, a valid path is a collision-free sequence of object poses and corresponding joint positions.

In Phase~$3$, where object pose estimation tends to be very noisy, the path is executed by simply tracking the planned joint positions with a PD controller. This approach does not rely on feedback of the object's pose, 
allowing us to be robust to inaccurate or missed object observations.  
In Phases~$1$ and $2$, however,  where pose estimations are more reliable, we also add feedback controllers: a force control policy similar to \citet{wuthrich2020trifinger} that helps maintain a stable grasp on the object, and a learned residual policy~\cite{Silver2018ResidualPL, Johannink19} that applies corrective position and torque actions to our controller.
In all phases, we combine this approach with a set of motion primitives that grasp and align the object before attempting to move it to the goal pose. Figure~\ref{fig:alg} provides an overview of our approach.

\subsection{Motion Primitives and Object Alignment}
\begin{figure}[!t]
    \centering
    \includegraphics[width=\linewidth]{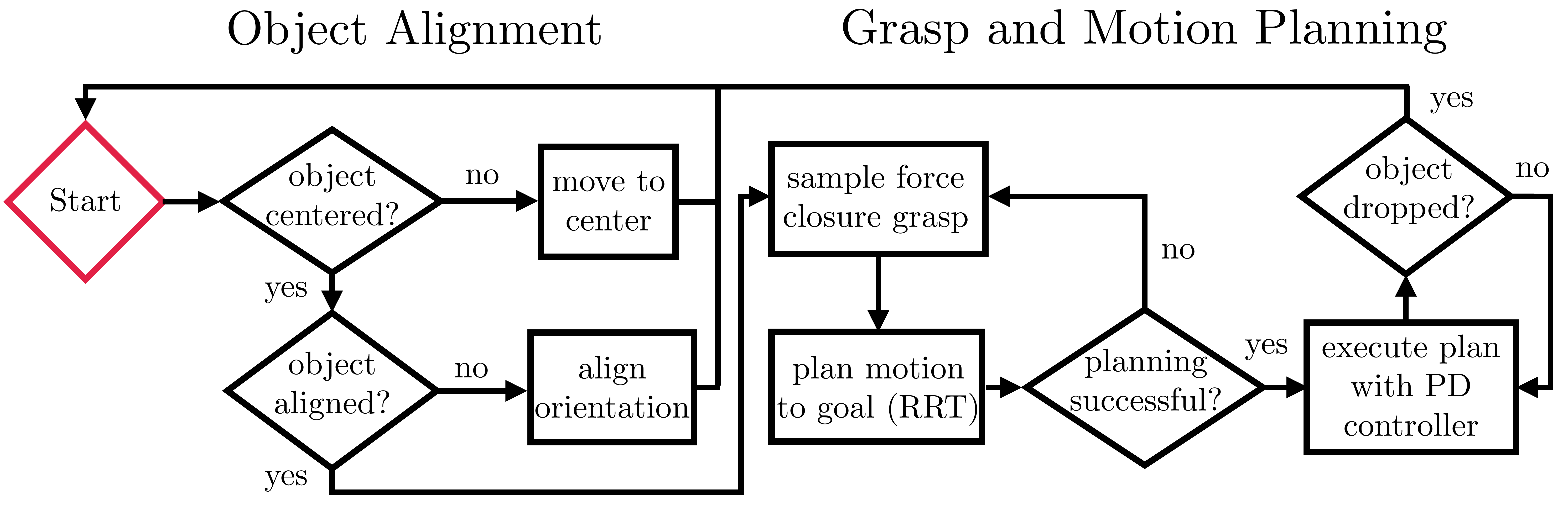}
    \caption{A visualization of the structure of our approach. We first align the pose of the object by moving it to the center of the workspace and 
    align the object with the goal orientation. We then find a grasp and subsequent motion plan to move the object to the goal pose. The object is then grasped and the plan is executed by tracking joint positions with a PD controller.}
    \label{fig:alg}
\end{figure}

Many goal poses cannot be reached with a single grasp. Consequently, we use a set of motion primitives to first align the object to the goal on the ground. These include: grasping the object, moving the object to the center of the workspace, and aligning the orientation of the object with the goal orientation.
These primitives are combined as shown in Figure~\ref{fig:alg}.

The \emph{grasping} motion primitive moves the fingertips in a straight line to a predefined pregrasp pose near the contact points, then grasps the object. This helps avoid accidentally hitting the object when approaching it. 
In Phase~$3$, the smallest sides of the object are about the same size as the fingertips. As a result, grasping is prone to failure if there is noise in the estimated pose of the object. To combat this, we first move the fingers to a position that does not occlude the different views of the object and collect several observations. We then average and project those observations onto the ground plane by setting the height of the object's position and projecting its orientation into the subspace of yaw-only rotations. We then use the averaged observations for grasping.

The \emph{move-to-center} motion primitive first grasps the object, then moves the fingertips along a straight-line path to the center of the workspace. In the case where the object is close to the edge of the workspace, inverse kinematics often fails to provide joint positions on some of the fingers. In such cases we pin those fingers at the default position, and execute the path only with available finger(s). 

To make it easier to find a grasp and subsequent plan that move the object to the goal, the \emph{align-yaw} and \emph{align-pitch} motion primitives rotate the object to be close to the goal orientation while remaining on the ground. For yaw rotations, this corresponds to projecting the goal-orientation onto the space of yaw-only rotations and rotating the object to the projected orientation. For pitch rotations, the pitch of the goal pose is rounded to the nearest $90$ degrees and the object is rotated to that orientation. These primitive are used only for Level~4 and only the align-pitch primitive is not used in Phase~$3$ because matching pitch does not affect the reward\footnotemark[1]. 
In Phases~$1$ and $2$, we used a predefined motion to rotate the object, but in Phase~$3$, we found that the small object size often lead to collisions between the fingers. Thus, in Phase~$3$, we use the grasp-planning loop described below to find a feasible grasp and subsequent plan that achieve the desired rotation. We then grasp the object, execute the planned motion, and release the object.


\subsection{Grasp-planning Loop}
\label{sec:grasp-planning-loop}
Once the object on the ground is reasonably centered and the orientation is aligned with the goal (only required at Level~$4$), the next step is to find a proper grasp and plan a path to the goal pose.
Since a goal pose is often only reachable with specific grasps, 
we repeat grasp sampling and path planning until a feasible path is found. We call this {\it grasp-planning loop}.

We first sample grasps. This is done in two ways: by randomly sampling a set of three contact points on the surface of the object, and by selecting from a set of predefined contact points. We then generate a grasp by assigning a fingertip to each contact point.
We reject grasps that do not exhibit the following two properties: the grasp is \textit{feasible}, meaning the finger tips can reach the contact points and there are no collisions between the fingers (excluding the tips) and the object, and the grasp is in \textit{force closure}, meaning we can resist any external force applied to the object. The force closure check is computed by assuming a Coulomb friction model and making a linear approximation to the friction cone.

Once a grasp has been sampled, we run an RRT~\cite{lavalle1998rapidly} in task space (i.e., the pose of the object) starting from the current object pose and planning to a desired object pose. We enforce that the grasp is maintained during planning by rejecting object poses if the corresponding grasp is not feasible. If no feasible path to the goal is found, we sample another grasp and repeat the process. To ensure we eventually find a feasible path, we slowly increase the size of the goal set of the RRT.

\subsection{Executing the plan}



Given a grasp and a path (sequence of joint positions and object poses) obtained from grasp-planning loop, we then follow the planned joint positions with a PD controller. During execution, we check to see if the object has been dropped. If so, we reset and plan again.
Notably, the execution of this plan does not rely on feedback of the object's pose. This provides robustness to errors in the provided estimates of the object's pose, including missed detections and occasional $180$~degree flips in the estimated orientation, which happens frequently in Phase~$3$.
In Phases~$1$ and $2$, without a notable issue on pose estimation, we additionally applied two feedback controllers: a force control policy and a learned residual policy. These policies are explained in detail in Section \ref{sec:feedback-controllers}

In cases in which the RRT generates a plan that does not move the object to the exact goal pose, we perform {\it Tip Adjustments} that subsequently append corrections to the plan after execution that attempt to move the object to the goal pose by translating and rotating the fingertip positions accordingly. In Phase~$2$, we also corrected for other errors such as slippage or changes in the grasp during execution of the plan based on estimated object pose, but we found the observation noise to be too large for this to work reliably in Phase~$3$.

\begin{table}
    \centering
    \setlength{\tabcolsep}{0em}
    \begin{tabularx}{\linewidth}{lcccc}
        \toprule
    {\bf Phase~1} & Level~$1$ & Level~$2$  & Level~$3$ & Level~$4$\\
    \midrule
    Force Control (FC) & $-138 \pm 143$ & $-532 \pm 402$ & $-458 \pm 477$ & $-1421 \pm 345$\\
    FC + Residual Learning (RL)& $\mathbf{-74 \pm 29}$ & $\mathbf{-123 \pm 69}\hphantom{0}$ & $\mathbf{-122 \pm 54\hphantom{0}}$ & $-1527 \pm 425$\\
    FC + Motion Planning (MP) & $-166 \pm 178$ & $-177 \pm 51\hphantom{0}$ & $-226 \pm 257$ & $-1127 \pm 509$\\
    FC + MP + RL & $-156 \pm 82\hphantom{0}$ & $-171 \pm 51\hphantom{0}$ & $-170 \pm 83\hphantom{0}$ & $\hphantom{0}\mathbf{-999 \pm 387}$\\
    \toprule
    {\bf Phase~2} \\
    \midrule
    FC & $-10051 \pm 5493$ & $-28817 \pm 4773\hphantom{0}$ & $-28414 \pm 8957\hphantom{0}$ & $-45320 \pm 5654\hphantom{0}$\\
    FC + RL& $-11192 \pm 6768$ & $-36405 \pm 7667\hphantom{0}$ & $-25324 \pm 11920$ & $-45370 \pm 13388$\\
    FC + MP & $\hphantom{0}-7973 \pm 7898\hphantom{0}$ & $-13559 \pm 9597\hphantom{0}$ & $-13842 \pm 12688$ & $-23517 \pm 10680$\\
    FC + MP + Tip Adjust.\ (TP) & $\hphantom{0}\mathbf{-2673 \pm 1166}$ & $\hphantom{0}-8115 \pm 8443\hphantom{-}$ & $\hphantom{0}-4830 \pm 5005\hphantom{-}$ & $-32572 \pm 15701$\\
    FC + MP + RL & $\hphantom{0}-8018 \pm 8011\hphantom{0}$ & $\hphantom{0}-7167 \pm 2170\hphantom{-}$ & $\hphantom{0}-8812 \pm 2996\hphantom{-}$ & $\mathbf{-20448 \pm 6154}\hphantom{0}$\\
    FC + MP + TP + RL & $\hphantom{0}-3330 \pm 4331\hphantom{0}$ & $\mathbf{\hphantom{0}-2877 \pm 204\hphantom{-0}}$ & $\mathbf{\hphantom{0}-3893 \pm 2453\hphantom{-}}$ & $-22301 \pm 11443$\\
    \toprule
    {\bf Phase~3} \\
    \midrule
    MP + TP & $-7944  \pm 4214$ & $\hphantom{0}-4039 \pm 446\hphantom{-0}$ & $-9896 \pm 10066$ & $-24755 \pm 10548$ \\
    \bottomrule
    \end{tabularx}
    \caption{Mean and standard deviation of rewards for different algorithms. We report the mean and standard deviation of rewards across $10$ episodes for each difficulty level. It should be noted that the specified frequency of control is largely different in Phase~1, and the magnitude of the reward differs from Phases~$2$ and 3 even though the evaluation metric is identical.}
    \label{tab:results}
\end{table}

\subsection{Feedback Controllers used in Phases $1$ and $2$}
\label{sec:feedback-controllers}
In Phases~$1$ and $2$, object pose estimation is reasonably accurate (Phase~$1$ presents no estimation errors since it is contained in a simulated environment).
Thus, we applied the following feedback control policies on top of position control: force control and residual policy.
Feedback controls enable the robot to exert auxiliary torque to maintain a stable grasp.

\subsubsection{Force Control}
Our force control pipeline is similar to the one described by \citet{wuthrich2020trifinger}. We use a PD controller to turn object pose errors into a wrench applied to the object's center of mass. Given a grasp of the object and its resulting contact points, we then solve for the tip forces necessary to achieve that wrench while maintaining friction cone constraints (see \citet{murray1994mathematical} for details). With tip forces, we combine Jacobian transpose control and gravity compensation via inverse dynamics to determine the joint torques. Force control was not used in Phase~$3$ due to noise in the provided pose estimation.

\subsubsection{Residual Policy Learning}
\label{sec:RL}
Residual policy learning~\cite{Silver2018ResidualPL, Johannink19} provides a simple way to improve a controller by learning corrective actions that are added to the output of the controller.
Additionally, learning corrections to a base controller can drastically simplify the reinforcement learning problem by removing the need for significant exploration outside
of the neighborhood around the base controller. It can also result in a policy that outperforms
both the base controller policy and a reinforcement learning policy trained from scratch.

We use Proximal Policy Optimization~\cite{schulman17} (PPO) to learn corrections to our controller in simulation. We fix the residual policy to zero for the first $1$\,M timesteps and only train the value function. This allows us to get a good estimate of the value of the base controller before we start learning. Additionally, we found that the initialization of the policy network was extremely important. Initializing the network to output action distributions with zero mean and small standard deviation greatly improved training as the combined controller stays close to the base controller early in training. Our reward function consists of three terms: the distance to goal, a regularizer on joint torques and velocities, and a penalty for the tips slipping on the object's surface. The last term encourages a firm grasp on the object, which is helpful for following the planned motion of the baseline controller.

While we planned to train models with domain randomization and real data, we did not have enough time. Instead, in Phase~$2$, we used the policies trained during Phase~$1$ and applied them directly to the real system. Surprisingly, despite substantial domain shift and changes to the base policies, we find the Phase~$1$ policies trained with our baseline motion planning controller were able to improve performance on the real robot. 
In Phase~$3$, we did not attempt any residual control.
We expect further improvements by applying sim-to-real methods \cite{tan18, peng18, tobin17} or fine-tuning with real data, and we leave this as a future work.
\section{Discussion}
\label{label:results}

Table \ref{tab:results} presents the performance of our approach across all phases along with several ablations. We find that our approach is able to consistently solve the tasks, is robust to observation noise, and can recover from failures (i.e., dropping the object). However, there is still room for improvement. We often find that executing our planned motions only takes the object near the goal pose. This is because the grasp was invalid at the exact goal pose, planning failed or took too long, or the object slipped while executing the plan. Robustly correcting for these errors could improve our performance considerably. Additionally we accumulate a lot of negative reward while the RRT is planning, and failures that trigger another round of planning can lead to a large variance in rewards. Using a Probabilistic Roadmap (PRM)~\cite{prm} would allow us to precompute many paths and thereby speed up planning significantly.

Due to time constraints, we were unable to try everything we had planned for Phase~$3$. Specifically, we implemented a framework for residual policy learning~\cite{Johannink19, Silver2018ResidualPL} with domain randomization~\cite{tobin17} to help transfer from simulation to the real robot, but did not have time to train and test the learned policies with a finalized controller. In Phases~$1$ and $2$, our policies were able to learn useful corrections to our controllers, and we believe such an approach would be beneficial in Phase~$3$ as well.

\section*{Acknowledgement}
We thank the organizers for all the effort they put into creating the TriFinger platform and hosting this competition. We are excited to use this platform to extend this work or for other research projects related to dexterous manipulation.

\bibliographystyle{plainnat}
{\small
\bibliography{references}
}

\end{document}